# A NOVEL PARALLEL RAY-CASTING ALGORITHM

YAN ZHANG, PENG GAO, XIAO-QING LI

Shenzhen Graduate School, Harbin Institute of Technology, China

**Abstract:**

The Ray-Casting algorithm is an important method for fast real-time surface display from 3D medical images. Based on Ray-Casting algorithm, a novel parallel Ray-Casting algorithm is proposed in this paper. A novel operation is introduced and defined as a star operation, and star operations can be computed in parallel in the proposed algorithm compared with the serial chain of star operations in the Ray-Casting algorithm. The computation complexity of the proposed algorithm is reduced from $O(n)$ to $O(\log_2^n)$.

## 1. Introduction

In recent years, the surface display from 3D medical image plays an important role in CT image display. Ray-Casting algorithm is the most widely used in the surface display from 3D medical image [1]. Ray-Casting algorithm can get more effective image information and have high value in clinical studies [2][3]. However, it is difficult to achieve real-time display by using the traditional Ray-Casting algorithm because the display speed is limited by the large number of radiation sampling and integration operations. Zhao et al. proposed an improved algorithm in 2015 [4], which changes the draw order of Ray-Casting, and puts the gradient estimation, classification, coloring and shading calculations in preprocessing stage to reduce transmission process computing tasks. Fan et al. proposed an efficient Ray-Casting volume algorithm in 2014 [5], which is improved from ray leaping with back steps. In the Fang Fun's algorithm, the technology of collision detection is introduced to reduce the number of projecting rays and avoid redundant ray sample calculations, and ray leaping method is used to skip empty voxel resampling in the collision detection bounding box and speed up the process of synthesis of ray, but a complex and inefficient pre-inspection process is used in the computer programming. Kang et al. proposed an algorithm in 2012 [6], which is based on CG language and programmed in GPU. In Kang's algorithm, a sampling function with early opacity termination is used to accelerate the speed of Ray-Casting.

Based on Ray-Casting algorithm, a novel parallel Ray-Casting is proposed in this paper. Parallel resampling and fusion are conducted in resampling stage of novel parallel Ray-Casting. The ray of novel parallel Ray-Casting be divided into a plurality of sections, serial resampling and fusion are calculated in each section in parallel way.

The paper is organized as follows. Section I is introduction. In section II, we introduce the mathematical background of Ray-Casting. In section III, a novel parallel Ray-Casting algorithm is proposed and described in detail. In section IV, the computation complexity and hardware resource are analyzed. Finally, section V gives the conclusion of the novel Ray Casting parallel algorithm.

## 2. Mathematical Background of Ray-Casting

The first step of Ray-Casting algorithm is data preparation. Then, resampling the data and two arrays are calculated, $C = (C_1, C_2, \cdots C_{i-1})$ and $O = (O_1, O_2, \cdots O_{i-1})$. Within the array C and O, $C_{i-1}$ and $O_{i-1}$ are the output color and opacity of the $(i-1)^{th}$ voxel respectively, and the input color and opacity of the $i^{th}$ voxel at the same time. Rays are then cast into these two arrays from the observer eyepoint. For $i^{th}$ voxel in each ray, a vector $(X_i, Y_i)$ represents the opacity and color at the $i^{th}$ voxel, $X_i$ and $Y_i$ is computed by tri-linearly interpolating from the colors and opacities in the 8 voxels closest to each sample location.

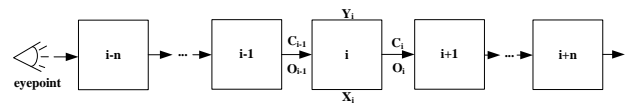

Figure 1. Schematic of merge process

Finally, assuming that the output color and opacity of the $(i-1)^{th}$ voxel are reached, the output color and opacity of the $i^{th}$ voxel can be calculated based on $C_{i-1}$, $O_{i-1}$, and the vector $(X_i, Y_i)$, as follows,

$$C_i = \frac{C_{i-1} \times O_{i-1} + Y_i \times X_i \times (1 - O_{i-1})}{O_i}$$
$$O_i = O_{i-1} + X_i \times (1 - O_{i-1}) \quad (1)$$

then

$$C_{i+1} = a_i \times C_i + b_i \quad (2)$$

where $a_i = \frac{O_{i-1}}{O_i}$, $b_i = \frac{Y_i \times X_i \times (1 - O_{i-1})}{O_i}$, $C_i = O_{i-1}$.

Based on the equation (2), the traditional Ray-Casting algorithm is a serial one because $C_{i+1}$ depends on $C_{i-1}$.

## 3. A Novel Algorithm

As shown in figure 1, the traditional Ray-Casting algorithm is serial. In order to meet the requirements of real-time performance of 3D Medical Image reconstruction, a novel parallel Ray Casting algorithm is proposed in this section, and a new operation is introduced.

Firstly, a star operation "*" is defined as,

$$(a_{[i,j]}, b_{[i,j]}) = (a_{[i,m]}, b_{[i,m]}) * (a_{[m-1,j]}, b_{[m-1,j]})$$
$$= (a_{[i,m]} a_{[m-1,j]}) + b_{[i,m]} \quad (i > m > j) \quad (3)$$

within $a_{[i,j]} = a_i a_{i-1} \cdots a_{j+1} a_j$,
$b_{[i,j]} = a_i a_{i-1} \cdots a_{j+1} b_j + a_i a_{i-1} \cdots a_{j+2} b_{j+1} + \cdots + a_{i-1} + b_j$.

Equation (2) leads to

$$C_{i+1} = a_i a_{i-1} \cdots a_2 a_1 c_{in} + a_i a_{i-1} \cdots a_2 b_1 +$$
$$a_i a_{i-1} \cdots a_3 b_2 + a_i a_{i-1} b_{i-2} + a_i b_{i-1} + b_i \quad (4)$$

Considering $C_{i+1}$ as a linear function of $c_{in}$, the coefficient of an item and constant term of equation (2) can be regarded as

$$[(a_i a_{i-1} \cdots a_2 a_1), (a_i a_{i-1} \cdots a_2 b_1 + a_i a_{i-1} \cdots a_3 b_2 +$$
$$a_i a_{i-1} b_{i-2} + a_i b_{i-1} + b_i)] \quad (5)$$

Above expression can be formalized as

$[(a_i a_{i-1} \cdots a_2 a_1), (a_i a_{i-1} \cdots a_2 b_1 + a_i a_{i-1} \cdots a_3 b_2 +$
$\qquad a_i a_{i-1} b_{i-2} + a_i b_{i-1} + b_i)]$
$= (a_i, b_i) * (a_{i-1} \cdots a_2 a_1, a_{i-1} \cdots a_2 b_1 + a_{i-1} \cdots a_3 b_2 + \cdots$
$\qquad + a_{i-1} b_{i-2} + b_{i-1})$
$= (a_i, b_i) * (a_{i-1}, b_{i-1}) * (a_{i-2} \cdots a_2 a_1, a_{i-2} \cdots a_2 b_1 +$
$\qquad a_{i-2} \cdots a_3 b_2 + \cdots + b_{i-2})$
$= \cdots\cdots$
$= (a_i, b_i) * (a_{i-1}, b_{i-1}) * \cdots * (a_2, b_2) * (a_1, b_1)$
$= (a_{[i,j]}, b_{[i,j]}) * (a_{[j-1,k]}, b_{[j-1,k]}) * \cdots * (a_{[m-1,n]}, b_{[m-1,n]}) *$
$\qquad (a_{[n-1,1]}, b_{[n-1,1]}) \qquad (i>j>k>\cdots>m>n>1) \quad (6)$

Obviously, equation (6) implies that input variables $(a_i, b_i)$ (or $(a_{[i,j]}, b_{[i,j]})$) meet associativity of "*" operation in traditional Ray Casting, so a novel parallel algorithm is reached. The equation is showed as,

$[(a_i a_{i-1} \cdots a_2 a_1), (a_i a_{i-1} \cdots a_2 b_1 + a_i a_{i-1} \cdots a_3 b_2 +$
$\qquad a_i a_{i-1} b_{i-2} + a_i b_{i-1} + b_i)]$
$= (a_1, b_1) * (a_2, b_2) * \cdots * (a_{i-1}, b_{i-1}) * (a_i, b_i)$
$= [(a_1, b_1) * (a_2, b_2)] * [(a_3, b_3) * (a_4, b_4)] * \cdots *$
$\qquad [(a_{i-3}, b_{i-3}) * (a_{i-2}, b_{i-2})] * [(a_{i-1}, b_{i-1}) * (a_i, b_i)]$
$= \cdots\cdots$
$= (a_{[1, n-1]}, b_{[1, n-1]}) * (a_{[n, m-1]}, b_{[n, m-1]}) * \cdots *$
$\qquad (a_{[k, j-1]}, b_{[k, j-1]}) * (a_{[j, i]}, b_{[j, i]})$
$\qquad (i>j>k>\cdots>m>n>1) \quad (7)$

## 4. Analysis of Computational Complexity

In figure 2 as follows, it is shown that the computation complexity is reduced from $O(n)$ to $O(\log_2^n)$ in the proposed parallel ray casting algorithm. Compared with only one-star operation unit used in the traditional ray casting algorithm, $\frac{n}{2}$ star operation units are need to be implemented in the proposed algorithm to achieve the computation complexity $O(\log_2^n)$.

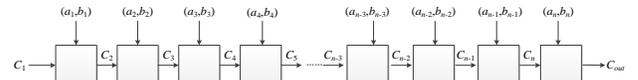

(a) Serial chain in the traditional ray casting algorithm

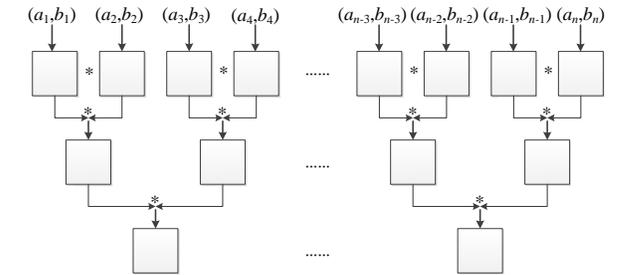

(b) binary tree structure in the parallel ray casting algorithm

Figure 2. Structure comparison between traditional and proposed casting algorithm

## 5. Conclusions

A novel parallel Ray-Casting algorithm is proposed based on traditional Ray-Casting algorithm in this paper. A new operation, which is called the star operation, is defined, and star operations can be computed in parallel since they meet the associative law. Compared with the traditional algorithms, the computation complexity of the proposed algorithm is reduced from $O(n)$ to $O(\log_2^n)$. In the parallel Ray-Casting algorithm, $\frac{n}{2}$ star operation units are needed to achieve the best performance.

## Acknowledgements

This work was supported in part by the technology development and innovation design fund of Nanshan, Shenzhen (Grant No. KC2013JSCX0041A), the scientific development fund of Shenzhen (Grant No. JC201005260168A), the Strategic emerging industry development fund of Shenzhen (Grant No. JCYJ20130329161328512), and the Science and Technology Planning Project of Guangdong Province (Grant No. 2013B090600105).

## References

[1] M. Levoy, "Display of Surface from Volume Data", IEEE Computer Graphics and Applications, Vol 8, No.5, pp. 29-37, 1988.
[2] T. Li, M. Xie, W. Zhao, and Y. Wei, "Shear-warp rendering algorithm based on radial basis functions interpolation", Proceeding of ICCMS10 Conference, IEEE, pp. 425-429, 2010.
[3] P. Lacroute, and M. Levoy, "Fast volume rendering using a shear-warp factorization of the viewing transformation", Proceeding of the 21st Annual Conference on Computer Graphics and Interactive Techniques, ACM, pp. 451-458, 1994.
[4] W. Zhao, T. Wan, T. Wu, and Y. Zhu, "Study on Improve the Ray-Casting Algorithm", Software Engineer, Vol 4, 2015.
[5] J. Fang, and X. Fang, "An efficient Ray-Casting Volume Rendering Algorithm", Computer Technology and Development, Vol 24, No. 8, pp. 67-70, 2014.
[6] J. Kang, B. Kang, J. Feng, et al, "Improved ray casting algorithm based on GPU programming", Computer Engineering and Applications, Vol 48, No. 1, pp. 199-201, 2012.